\documentclass[11pt,a4paper]{article}
\usepackage[hyperref]{emnlp2020}
\usepackage{times}
\usepackage{latexsym}
\usepackage{graphicx}
\usepackage{booktabs}

\usepackage{microtype}

\usepackage{amsmath}
\usepackage{adjustbox}
\usepackage{multirow}

\aclfinalcopy 
\def\aclpaperid{***} 

\newcommand{\ignore}[1]{}

\title{
Understanding the Extent to which Summarization Evaluation Metrics Measure the Information Quality of Summaries
}

\author{Daniel Deutsch and Dan Roth \\
  Department of Computer and Information Science \\
  University of Pennsylvania \\
  \texttt{\{ddeutsch,danroth\}@seas.upenn.edu}\\}

\date{}

\begin{document}

\maketitle

\begin{abstract}
Reference-based metrics such as ROUGE \citep{Lin2004} or BERTScore \citep{Zhang2019} evaluate the content quality of a summary by comparing the summary to a reference.
Ideally, this comparison should measure the summary's information quality by calculating how much information the summaries have in common.
In this work, we analyze the token alignments used by ROUGE and BERTScore to compare summaries and argue that their scores largely cannot be interpreted as measuring information overlap, but rather the extent to which they discuss the same topics.
Further, we provide evidence that this result holds true for many other summarization evaluation metrics.
The consequence of this result is that it means the summarization community has not yet found a reliable automatic metric that aligns with its research goal, to generate summaries with high-quality information.
Then, we propose a simple and interpretable method of evaluating summaries which does directly measure information overlap and demonstrate how it can be used to gain insights into model behavior that could not be provided by other methods alone.\footnote{
    Our code will be available at \url{https://github.com/CogComp/content-analysis-experiments}.
}
\end{abstract}
\section{Introduction}
The development of a reliable metric that can automatically evaluate the content of a summary has been an active area of research for nearly two decades.
Over the years, many different metrics have been proposed \citep[][\emph{i.a.}]{Lin2004, Hovy2006,Giannakopoulos2008,Louis2013,Zhao2019,Zhang2019}. 

\begin{figure}
    \centering
    \includegraphics{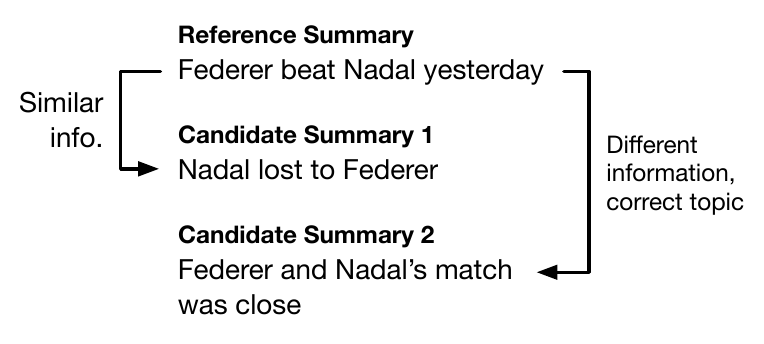}
    \caption{
    Both candidate summaries are similar to the reference, but along different dimensions: Candidate 1 contains some of the same information, whereas candidate 2's information is different, but it is at least discusses the correct topic.
    The goal of this work is to understand if summarization evaluation metrics' scores should be interpreted as measures of information overlap or, less desirably, topic similarity.
    }
    \label{fig:intro_fig}
\end{figure}

The majority of these of approaches score a candidate summary by measuring its similarity to a reference summary.
The most popular metric, ROUGE \citep{Lin2004}, compares summaries based on their lexical overlap, whereas more recent methods, such as BERTScore \citep{Zhang2019}, do so based on the similarity of the summary tokens' contextualized word embeddings.

Ideally, metrics that evaluate the content of a summary should measure the quality of the information in the summary.
However, it is not clear whether metrics such as ROUGE and BERTScore evaluate summaries based on how much information they have in common with the reference or some less desirable dimension of similarity, such as whether the two summaries discuss the same topics (see Fig.~\ref{fig:intro_fig}).

In this work, we demonstrate that ROUGE and BERTScore largely do not measure how much information two summaries have in common.
Further, we provide evidence that suggests this result holds true for many other evaluation metrics as well.
Together, these point to a grim conclusion for summarization evaluation: the community has not yet found an automatic metric that successfully evaluates the quality of information in a summary.

Our analysis casts ROUGE and BERTScore into a unified framework in which the similarity of two summaries is calculated based on an alignment between the summaries' tokens (\S\ref{sec:framework}).
This alignment-based view of the metrics enables performing two different analyses of how well they measure information overlap.

The first analysis demonstrates that only a small portion of the metrics' token alignments are between phrases which contain identical information according to domain experts (\S\ref{sec:scu_analysis}).
The second reveals that token alignments which represent common information are vastly outnumbered by those which represent the summaries discussing the same topic (\S\ref{sec:category_analysis}).
Overall, both analyses support the conclusion that ROUGE and BERTScore largely do not measure information overlap.

Then, we expand our analysis to consider if 10 other evaluation metrics successfully measure information quality or not (\S\ref{sec:other_metrics}).
The experiments show that all 10 metrics correlate more strongly to ROUGE than to the gold-standard method of manually comparing two summaries' information, the Pyramid Method \citep{Nenkova2004}, and that each correlates to the Pyramid Method around the same level as ROUGE.
This is evidence that the other metrics measure information overlap no better than ROUGE does.

Finally, we propose a simple and interpretable evaluation method which \emph{does} directly measure information overlap by calculating metrics on how many tuples of tokens that represent information are common between summaries (\S\ref{sec:fine_grained_comparison}).
Using our methodology, we are able to show that state-of-the-art summarization systems do produce summaries with better content than baseline models, an insight that ROUGE and BERTScore alone could not provide.

The contributions of this work include (1) an analysis which reveals that ROUGE and BERTScore largely do not measure the information overlap between two summaries, (2) evidence that many other evaluation metrics likely suffer from the problem, and (3) a proposal of a simple and interpretable evaluation method which does directly measure information overlap.

\section{Understanding Evaluation Metrics}
Reference-based evaluation metrics assume that human-written reference summaries have gold-standard content and score a candidate summary based on its similarity to the reference.
An ideal evaluation metric that measures the content quality of a summary should score the quality of its \emph{information}.
For reference-based metrics, this means that the comparison between the two summaries should measure how much information they have in common.

Metrics such as ROUGE and BERTScore calculate the similarity of two summaries either by how much lexical overlap they have or how similar the summaries' contextual word embeddings are (discussed in more detail in \S\ref{sec:framework}).
Although we understand how their scores are calculated, it is not clear how the scores should be interpreted: Are they representative of how much information the two summaries have in common, or do they describe how similar the summaries are on some other less desirable dimension, such as whether they discuss the same topics?
Our goal is to answer this question.

Knowing the answer is critically important.
The goal of summarization is to produce summaries which contain the ``correct'' information (among other desiderata).
Automatic metrics are the most frequent method that researchers use to argue that one summarization model generates better summaries than another.
If our evaluation metrics are not aligned with our research goals --- or if we do not understand what they measure at all --- then we do not know whether we are making progress as a community.

\section{A Common Framework}
\label{sec:framework}
The focus of our analysis will be primarily on two evaluation metrics, ROUGE and BERTScore.
Although on the surface these two metrics appear to compare two summaries very differently, here we demonstrate how they can both be viewed as calculating a score based on a weighted alignment between the summaries' tokens.
This common framework enables us to reason about how to interpret their scores.

Let $R = r_1, r_2, \dots, r_m$  and $S = s_1, s_2, \dots, s_n$ be the tokens of the reference and candidate summaries.
ROUGE-1 counts the number of unigrams that are in common between the two summaries:\footnote{
    Our analysis focuses on the unigram variant of ROUGE, called ROUGE-1.
    We refer to it, where clear, as ROUGE for simplicity.
}
\begin{equation}
    M = \sum_{s \in S} \min(c_R(s), c_S(s))
\end{equation}
where $c(s)$ counts the number of times $s$ appears in the respective summaries and the summand is over unique unigrams.
Then, precision and recall are calculated by dividing $M$ by $n$ and $m$, respectively.
When multiple references are available, the precision and recall scores are micro-averaged.

A weighted alignment $A$ is a set of token alignments $(i, j, w)$ that map token $r_i$ to $s_j$ with weight $w \in (0, 1.0]$.
The weight of an alignment, denoted $W(A)$, is the sum of the weights of the individual token alignments.
ROUGE can be viewed as creating an alignment by pairing $\min(c_R(s), c_S(s))$ occurrences of unigram $s$ in $R$ and $S$ with weight $1.0$ for all unigrams.
It additionally imposes a constraint that each token can be aligned to at most one other token.
Since a unigram may appear multiple times in a summary, the alignment may not be unique, however, its weight will equal $M$.

BERTScore calculates a similarity score between two pieces of text based on the pairwise cosine similarities of their tokens' BERT embeddings.
Let $B_{ij}$ be the similarity score between the embeddings for $r_i$ and $s_j$.
To calculate recall, BERTScore first aligns every reference token to its most-similar summary token (Eq.~\ref{eqn:bertscore_alignment}).
Then, the sum of the corresponding similarities is normalized by the number of reference tokens to get the recall score (Eq.~\ref{eqn:bertscore_recall}).
\begin{equation}
    \label{eqn:bertscore_alignment}
    A_R = \{(i, j, B_{ij}): \forall i, j = \arg\max_k B_{ik}\}
\end{equation}
\begin{equation}
    \label{eqn:bertscore_recall}
    \textrm{BERTScore}_{\textrm{Recall}} = \frac{W(A_R)}{m}
\end{equation}

A similar procedure is followed to calculate precision, but instead, every summary token is aligned to its most-similar reference token, and the sum of the similarities is normalized by the number of summary tokens.
When multiple references are available, the precision and recall scores are defined to be the maximum respective values across references.

By formulating ROUGE and BERTScore in a framework based on token alignments, we can reason about their behaviors by examining the tokens they align in two different analyses, as described next.

\section{SCU-Based Analysis}
\label{sec:scu_analysis}
The first analysis compares the two metrics' token alignments to annotations derived from the Pyramid Method \citep{Nenkova2004}.

The Pyramid Method is a technique to manually evaluate the content of a candidate summary by comparing it to a set of reference summaries.
The method uses a domain-expert annotator to exhaustively identify atomic units of meaning in the summaries, known as summary content units (SCUs), and mark their occurrences in the reference and candidate summaries.
Two phrases marked with the same SCU are considered to express the same information.
Since the Pyramid Method annotation is exhaustive, we can assume that any two phrases in the reference and candidate summaries that are not marked with the same SCU do not have the same meaning.

These annotated phrases can be used to reason about ROUGE and BERTScore:
If a large portion of their token alignments is between phrases that express the same information, then their scores can potentially be interpreted as representing the summaries' information overlap.
Otherwise, it is evidence that they do not compare summaries based on information.

\begin{figure}
    \centering
    \includegraphics{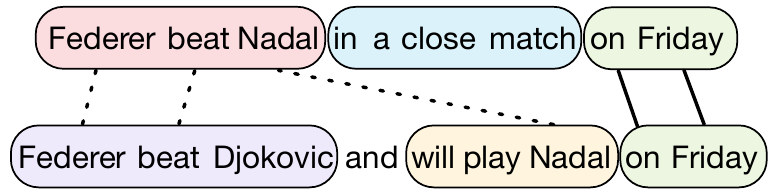}
    \caption{
    An example token alignment used by ROUGE or BERTScore.
    Each color represents a summary content unit (SCU) that marks informational content.
    Only $2/5$ of the token alignments (the solid edges) can be explained by matches between phrases that express the same information (the green phrases).
    }
    \label{fig:scu_example}
\end{figure}

For this analysis, we use the summaries and Pyramid annotations from the TAC 2008 and 2009 multi-document summarization datasets \citep{Dang2008,Dang2009}.
These datasets have 48/44 clusters of around 10 documents, each with 4 reference summaries and 58/55 system summaries that have been annotated with SCUs, respectively.

For each of the system summaries, we calculate the proportion of the total alignment weight that can be explained by matches between identical SCUs, as defined in Eqs.~\ref{eqn:a_scu} and \ref{eqn:scu_prop}:
\begin{equation}
    \label{eqn:a_scu}
    \begin{split}
        A_\textrm{SCU} = \{(i, j, w): \;& (i, j, w) \in A, \\
        &\textrm{SCU}(i) \cap \textrm{SCU}(j) \neq \emptyset\}
    \end{split}
\end{equation}
\begin{equation}
    \label{eqn:scu_prop}
    \textrm{Prop}_\textrm{SCU} = \frac{W(A_\textrm{SCU})}{W(A)}
\end{equation}
where $\textrm{SCU}(i)$ returns the set of SCUs that are annotated for the token at index $i$.
Fig.~\ref{fig:scu_example} has an example of this calculation.
Since ROUGE does not use a unique alignment, we choose the alignment which maximizes Eq.~\ref{eqn:scu_prop}, thus calculating an upper-bound.

\begin{figure}
    \centering
\includegraphics[width=\columnwidth]{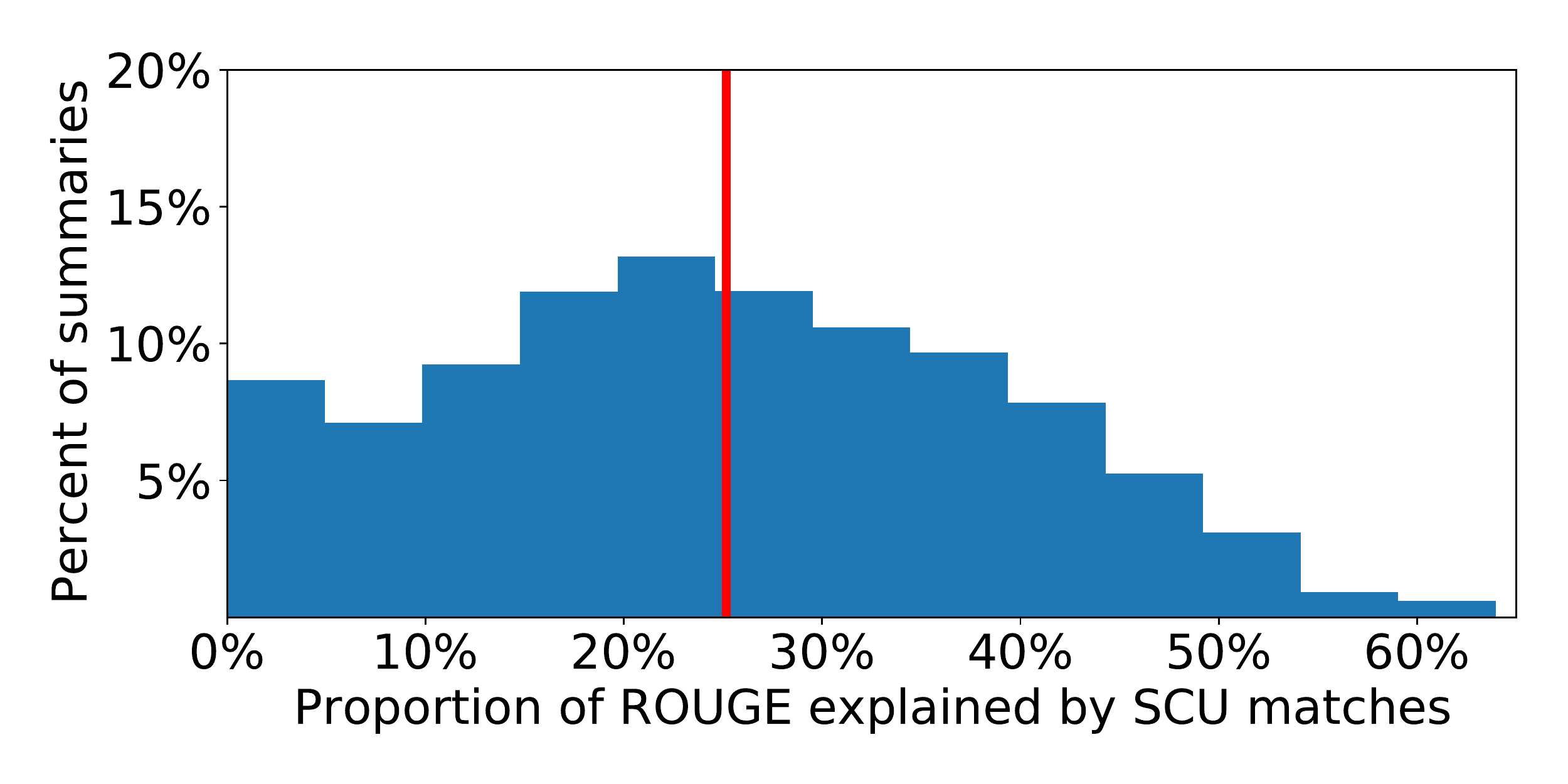}
\includegraphics[width=\columnwidth]{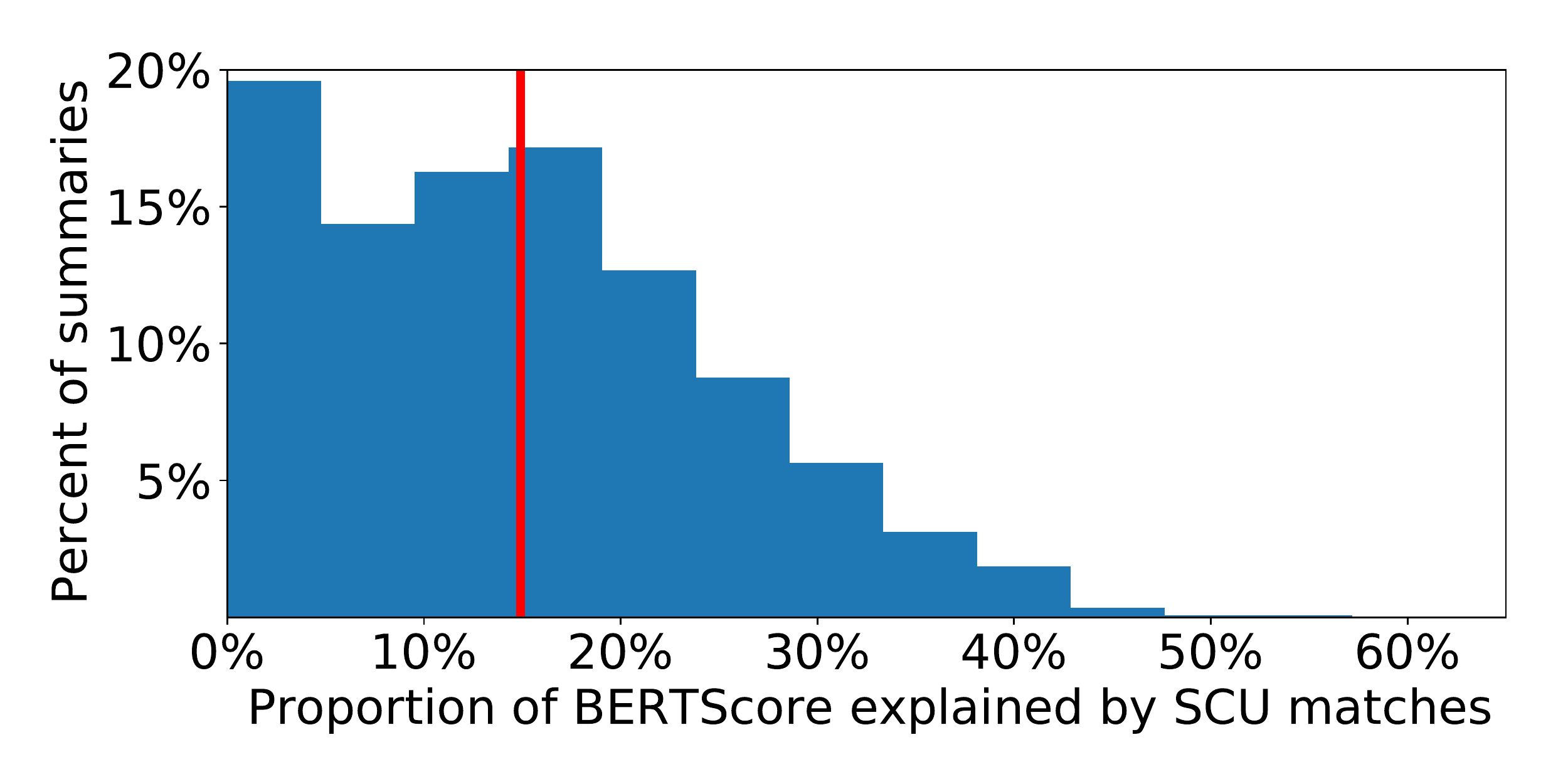}
    \caption{The distribution of the proportion of ROUGE (top) and BERTScore (bottom) on TAC 2008 that can be explained by tokens matches that are labeled with the same SCU (Eq.~\ref{eqn:scu_prop}).
    The averages, 25\% and 15\% (in red), indicate that only a small amount of their scores is between phrases that express the same information.
    }
    \label{fig:scu-comparison}
\end{figure}

The distribution of the proportion of ROUGE and BERTScore explained by SCU matches is presented in Figure~\ref{fig:scu-comparison}.
We find that, on average, only 25\% and 15\% of these metrics scores comes from matches between tokens marked with the same SCUs.
Since only a relatively small fraction of the overall metric scores comes from phrases with the same information, this suggests that ROUGE and BERTScore's values cannot be interpreted as a measure of information overlap.

\section{Category-Based Analysis}
\label{sec:category_analysis}
The second analysis of ROUGE and BERTScore focuses on grouping token alignments into categories (\S\ref{sec:categories}), then using those categories to reason about how much of the metrics' scores is explained by information or topic matches (\S\ref{sec:category_analysis}).

\subsection{Token Alignment Categorization}
\label{sec:categories}
We define a category to be a function $C$ that selects the subset of summary token indices for which that category applies.
For example, a ``noun'' category would select only the token indices that correspond to nouns.
$C(S)$ denotes the application of a category to summary $S$.

Each category is used to filter an alignment $A$ used by ROUGE or BERTScore to a category-specific alignment between tokens which belong to that category only, denoted $A_C$:
\begin{equation}
    \begin{split}
        A_C = \{(i,j,w) :\;& (i,j,w) \in A, \\
            &i \in C(R), j \in C(S)\}
    \end{split}
\end{equation}
For the ``noun'' category, $A_C$ would be the subset of token alignments between nouns in $R$ and $S$.

\begin{figure}
    \centering
    \includegraphics{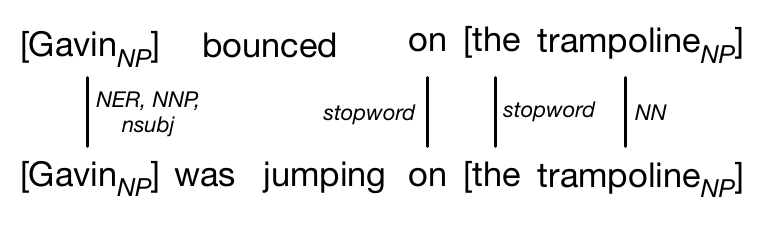}
    \caption{
    Every token alignment used by ROUGE or BERTScore is assigned to one or more interpretable categories (defined in \S\ref{sec:category_analysis}).
    This allows us to calculate, for this example, that matches between named-entities contribute $1/4$ to the overall score, stopwords $2/4$, and noun phrases $3/4$ (assuming alignment weights of $1.0$).
    }
    \label{fig:category_example}
\end{figure}

Then, the \emph{contribution} of $C$ is defined as the ratio between $A_C$ and $A$:
\begin{equation}
    \label{eqn:contribution}
    \textrm{Contribution}_C = \frac{W(A_C)}{W(A)}
\end{equation}
The contribution of $C$ can be interpreted as the portion of ROUGE or BERTScore that can be explained by matches between tokens in category $C$ (see Fig.~\ref{fig:category_example} for an example).

\paragraph{Higher-Order Categories}
Although our analysis only uses unigram alignments, it is desirable to reason about groups of tokens.
This would enable calculating how much of the metrics' scores can be explained by matches between (subject, verb, object) tuples, for instance.

\begin{figure}[t]
    \centering
    \includegraphics[width=0.8\columnwidth]{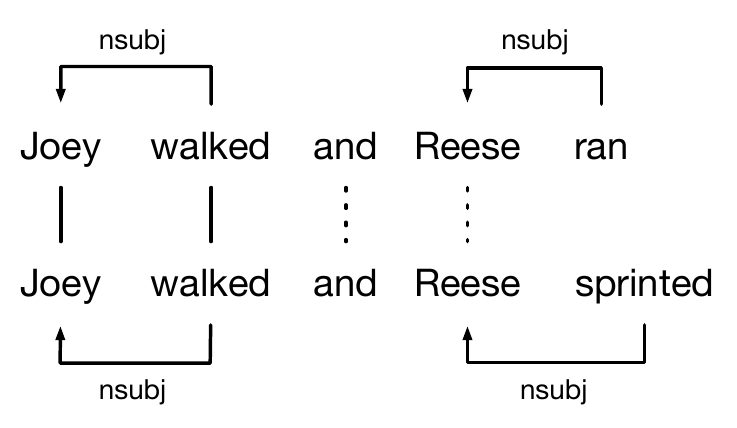}
    \caption{
    The \textsc{vb+nsubj} category selects tuples of verbs and their corresponding \textsc{nsubj} dependents in the dependency tree.
    In this example, $2/4$ of the alignment (the solid lines) can be explained by matches between such tuples.
    The dashed lines cannot:
    The ``and'' alignment is not part of any tuple;
    Since ``ran'' and ``sprinted'' are not aligned, their corresponding tuples are not considered to be aligned, so the ``Reese'' match does not count toward the total.
    }
    \label{fig:tuple-example}
\end{figure}

We extend the definition of a category to select a set of \emph{tuples} of indices.
Then $A_C$ selects only the token alignments in $A$ that are included in an aligned tuple selected by $C$.
Two tuples $(i_1, \dots, i_k)$ and $(j_1, \dots, j_k)$ are said to be aligned if indices $i_\ell$ and $j_\ell$ are aligned for $\ell = 1, \dots, k$.
Fig.~\ref{fig:tuple-example} has an example tuple-based matching.


\subsection{Category-Based Analysis}
\label{sec:category_analysis}
Next, we define a set of categories in which each category represents either information or topic matches, then reason about how much of the metrics' scores can be explained by information or topic similarities based on the corresponding category contributions.

We define the following categories:
\begin{enumerate}
    \item \textbf{Stopwords}: One category to select matches between stopwords, denoted \textsc{stopwords}.

    \item \textbf{Parts-of-Speech}: Six categories, one for selecting alignments between each type of the following part-of-speech tags: common nouns (\textsc{nn}), proper nouns (\textsc{nnp}), verbs (\textsc{vb}), adjectives (\textsc{adj}), adverbs (\textsc{adv}), and numerals (\textsc{num}).
    
    \item \textbf{Named-Entity}: One category for all named-entities, denoted \textsc{ner}.
    This category only selects alignments between tokens if they are the same type of named-entity (person, location, or organization).
    
    \item \textbf{NP Chunks}: One category to select matches between tokens that are part of noun phrases, denoted \textsc{np-chunks}.
    
    \item \textbf{Dependency}: Three categories that select matches between tokens with the same dependency tree arc label for \textsc{root}, \textsc{nsubj}, and \textsc{dobj} labels.
    
    \item \textbf{Dependency Tuples}: Three categories that match higher-order tuples based on the dependency tree.
    Each category selects a tuple containg a verb and either its subject child (\textsc{nsubj}), object child (\textsc{dobj}), or both.
    These categories are denoted \textsc{vb+nsubj}, \textsc{vb+dobj}, and \textsc{vb+nsubj+dobj}.
    They are representative on information expressed as (subject, verb, object) tuples, for instance.
    
\end{enumerate}
We consider 2 through 5 to be keywords that represent the topics discussed in the summaries, whereas 6 describes tuples which express the summaries' information.

\begin{table}[t]
    \centering
    \begin{adjustbox}{width=\columnwidth}
    \begin{tabular}{cccccc}
        \toprule
         & \multicolumn{2}{c}{TAC'08} & &  \multicolumn{2}{c}{CNN/DM} \\
        \cmidrule{2-3} \cmidrule{5-6}
        Category & R & BS & & R & BS \\
        \midrule
        \textsc{np-chunks}     & 58.7 & 46.1 & & 53.6 & 43.0 \\
        \textsc{stopwords}     & 54.6 & 32.4 & & 48.4 & 28.7 \\
        \textsc{nn}            & 17.9 & 13.7 & & 31.8 & 24.9 \\
        \textsc{nnp}           & 14.9 & 11.3 & & 0.3  & 0.2  \\
        \textsc{ner}           & 13.5 & 8.5  & & 0.1  & 0.1  \\
        \textsc{vb}            & 9.0  & 9.3  & & 14.1 & 10.6 \\
        \textsc{adj}           & 4.1  & 2.6  & & 6.2  & 4.0  \\
        \textsc{nsubj}         & 3.9  & 2.2  & & 6.3  & 4.1  \\
        \textsc{dobj}          & 2.0  & 1.4  & & 2.8  & 1.7  \\
        \textsc{num}           & 1.5  & 1.7  & & 2.5  & 1.9  \\
        \textsc{vb+dobj}       & 1.3  & 0.4  & & 3.4  & 1.0  \\
        \textsc{root}          & 1.1  & 1.5  & & 3.3  & 2.5  \\
        \textsc{vb+nsubj}      & 1.0  & 0.5  & & 3.8  & 2.4  \\
        \textsc{adv}           & 0.6  & 0.4  & & 1.6  & 0.8  \\
        \textsc{vb+nsubj+dobj} & 0.3  & 0.1  & & 1.5  & 0.5  \\
        \bottomrule
    \end{tabular}
    \end{adjustbox}
    \caption{The contributions (Eq.~\ref{eqn:contribution}) of every category to ROUGE (R) and BERTScore (BS) on TAC 2008 and CNN/DailyMail indicate the metrics are largely matching nouns and stopwords rather than tuples which express information (e.g., \textsc{vb+nsubj+dobj}).
    The contributions do not sum to 100\% because more than one category can explain the same token alignment.
    The \textsc{nnp} and \textsc{ner} for CNN/DailyMail are significantly lower because the candidate summaries were all lower-cased.
    }
    \label{tab:contributions}
\end{table}

The contributions of each category on the TAC 2008 summaries as well as the summaries produced by baseline \citep{See2017} and state-of-the-art \citep{Liu2019} abstractive models on the CNN/DailyMail dataset \citep{Nallapati2016} is presented in Table~\ref{tab:contributions}.

The results across datasets and evaluation metrics largely follow the same trend:
Noun- and stopword-based matches explain the vast majority of the token alignments used by both ROUGE and BERTScore, whereas the dependency tuple categories explain very little of the overall scores.\footnote{
    Although there are versions of ROUGE that remove stopwords, including them is significantly more common, and therefore we analyze the more popular ROUGE variant.
}
For instance, on TAC 2008, noun phrase and stopword matches contribute 58.7\% and 54.6\% to ROUGE, whereas the dependency tuple with the largest contribution, \textsc{vb+dobj} only contributes 1.3\%.

When the specific categories are grouped by content type in Table~\ref{tab:content-contributions}, it becomes even more apparent that topic and stopwords matches explain most of ROUGE and BERTScore.\footnote{
    The numbers in Table~\ref{tab:content-contributions} numbers cannot be directly read off Table~\ref{tab:contributions} nor do they sum to 100\% because multiple categories can explain the same token alignment.
}
Specifically, we find that topic, stopword, and information matches explain 70.1\%, 54.6\%, and 2.2\% of ROUGE's score on TAC 2008.

The low contribution of information-based categories toward each metric is further evidence that neither metric strongly captures the information overlap between summaries, supporting the results found in \S\ref{sec:scu_analysis}.

\begin{table}
    \centering
    \begin{tabular}{cccccc}
        \toprule
         & \multicolumn{2}{c}{TAC'08} & &  \multicolumn{2}{c}{CNN/DM} \\
        \cmidrule{2-3} \cmidrule{5-6}
        Content Type & R & BS & & R & BS \\
        \midrule
        Topic        & 70.6 & 57.9 & & 75.0 & 59.2 \\
        Information  & 2.2  & 0.9  & & 6.7  & 3.2  \\
        Stopwords    & 54.6 & 32.4 & & 48.4 & 28.7 \\
        \bottomrule
    \end{tabular}
    \caption{The contributions of different categories of token matches when grouped by whether they represent topics, information, or stopwords.
    Clearly, the information categories explain only a small proportion of the overall metrics scores on TAC'08 and CNN/DailyMail.
    }
    \label{tab:content-contributions}
\end{table}

\begin{table}[h]
    \centering
    \begin{adjustbox}{width=\columnwidth}
    \begin{tabular}{ccccc}
        \toprule
        Category & PG & BSEA & $\Delta$ & Rel. $\Delta$ \\
        \midrule
        ROUGE & 39.2 & 41.9 & 2.7 & 6.9 \\
        \midrule
        
        \textsc{vb+nsubj+dobj} & 5.8 & 7.1 & 1.3 & 22.4 \\
        \textsc{root} & 19.1 & 23.3 & 4.2 & 22.0 \\
        \textsc{num} & 22.0 & 26.6 & 4.6 & 20.9 \\
        \textsc{vb+nsubj} & 10.0 & 12.0 & 2.0 & 20.0 \\
        \textsc{adv} & 13.8 & 16.3 & 2.5 & 18.1 \\
        \textsc{vb+dobj} & 12.7 & 14.6 & 1.9 & 15.0 \\
        \textsc{dobj} & 18.7 & 20.9 & 2.2 & 11.8 \\
        \textsc{vb} & 28.9 & 32.1 & 3.2 & 11.1 \\
        \textsc{adj} & 26.5 & 29.1 & 2.6 & 9.8 \\
        \textsc{nsubj} & 29.5 & 32.2 & 2.7 & 9.2 \\
        \textsc{nn} & 38.0 & 40.9 & 2.9 & 7.6 \\
        \textsc{np-chunks} & 39.0 & 41.6 & 2.6 & 6.7 \\
        \textsc{stopwords} & 42.0 & 44.1 & 2.1 & 5.0 \\
        \textsc{nnp} & 5.0 & 5.0 & 0.0 & 0.0 \\
        \textsc{ner} & 1.4 & 1.1 & -0.3 & -21.4 \\
        \bottomrule
    \end{tabular}
    \end{adjustbox}
    \caption{The absolute and relative category-specific F$_1$ (Eqs.~\ref{eqn:category_recall} and \ref{eqn:category_precision}) differences between the Pointer-Generator model (PG) and the BertSumExtAbs model (BSEA) demonstrate that the state-of-the-art model does generate information that is more similar to the ground-truth (e.g., changes in \textsc{vb+nsubj+dobj}), but there is still significant room for improvement (low corresponding F$_1$ scores).
    }
    \label{tab:ccndm-comparison}
\end{table}
\section{Other Evaluation Metrics}
\label{sec:other_metrics}
The analyses thus far have exploited the structure of ROUGE and BERTScore to reason about the extent to which they measure information overlap between two summaries.
Although it is desirable to ask the same question about other evaluation metrics, the metrics may not directly fit into this analysis framework or it would require significant effort to repeat this analysis for each one.
Instead, we indirectly reason about how much information overlap other metrics measure through their correlations to ROUGE and the Pyramid Method as follows.

First, we assume that the Pyramid Score is the gold-standard for measuring the information overlap between summaries.
This is a relatively safe assumption because the Pyramid Method is annotated by domain experts, and a candidate's Pyramid Score is based solely on how much information it has in common with a reference.
There is no credit given to a candidate for discussing the right topics but with the incorrect information.

Then, the correlations of the other metrics to both ROUGE and the Pyramid Score are calculated and compared.
If the correlation to ROUGE is higher than the correlation to the Pyramid Score, then it is more likely that the metric suffers from the same issues that ROUGE does than it is to directly measure information overlap.

Table~\ref{tab:other_metrics} contains the summary-level correlations of various other evaluation metrics to ROUGE and the Pyramid Score.
The other metrics are: AutoSummENG \citep{Giannakopoulos2008}, BEwT-E \citep{Tratz2008}, MeMoG \citep{Giannakopoulos2011}, METEOR \citep{Denkowski2014}, MoverScore \citep{Zhao2019}, NPowER \citep{Giannakopoulos2013}, PyrEval \citep{Gao2019}, ROUGE-2, S$^3$ \citep{Peyrard2017}, and SUPERT \citep{Gao2020}.
These metrics exhibit a variety of different comparison techniques, from $n$-gram graph comparisons to contextual word-embedding comparisons and other alignment based approaches.

\begin{table}[]
    \centering
    \begin{adjustbox}{width=\columnwidth}
    \begin{tabular}{cccc}
        \toprule
        Metric & ROUGE-1 & Pyr. Score & $\Delta$ \\
        \midrule
        ROUGE-1 & 1.00 & 0.59 & - \\
        Pyramid Score & 0.59 & 1.00 & - \\
        \midrule
        AutoSummENG & 0.83 & 0.61 & 0.22 \\
        BERTScore & 0.74 & 0.59 & 0.16 \\
        BEwT-E & 0.81 & 0.62 & 0.19 \\
        MeMoG & 0.68 & 0.52 & 0.16 \\
        METEOR & 0.91 & 0.63 & 0.28 \\
        MoverScore & 0.79 & 0.61 & 0.18 \\
        NPowER & 0.81 & 0.60 & 0.21 \\
        PyrEval & 0.47 & 0.35 & 0.12 \\
        ROUGE-2 & 0.79 & 0.58 & 0.21 \\
        S$^3$ & 0.92 & 0.63 & 0.29 \\
        SUPERT & 0.55 & 0.49 & 0.06 \\
        \bottomrule
    \end{tabular}
    \end{adjustbox}
    \caption{The summary-level Pearson correlations of various metrics to ROUGE-1 and the Pyramid Score ($\Delta$ is the difference between them).
    All of the other metrics correlate more strongly to ROUGE-1 than the Pyramid Score (by around 0.2) and correlate to the Pyramid Score approximately as much as ROUGE-1 does (around 0.6).
    Together, these results suggest the other metrics are as poor evaluators of information overlap as ROUGE-1 is.
    }
    \label{tab:other_metrics}
\end{table}

Notably, all of the metrics' Pearson correlations to ROUGE are higher than to the Pyramid Score, generally by around 0.2 points, suggesting these metrics do not measure information overlap well.
Further, their correlations to the Pyramid Score are roughly the same as ROUGE's, around 0.6.
This is means that these metrics correlate to a direct measure of information overlap as well as one would expect a metric which measures information overlap at the level of ROUGE to correlate.
Although the results of this experiment are not direct evidence that the other evaluation metrics do a poor job at measuring information overlap, they do strongly suggest it.

An ideal content evaluation metric should measure how much information two summaries have in common, and overall, this experiment indicates that the summarization community has not yet found any such automatic metric.

\section{An Interpretable System Comparison}
\label{sec:fine_grained_comparison}
Thus far, we have argued that current evaluation metrics do not strongly measure how much information two summaries have in common.
In this Section, we aim to propose a simple and interpretable method of evaluating summaries that \emph{does} directly measure information overlap.

Instead of evaluating summarization systems with ROUGE or BERTScore, we instead calculate a set of precision and recall scores based on the categories defined in \S\ref{sec:category_analysis}.
We define the category-specific precision and recall as:
\begin{align}
    \label{eqn:category_recall}
    \textrm{Recall}_C &= \frac{W(A_C)}{\vert C(R)\vert} \\
    \label{eqn:category_precision}
    \textrm{Precision}_C &= \frac{W(A_C)}{\vert C(S)\vert}
\end{align}
These can be understood as calculating ROUGE or BERTScore on the subset of tokens defined by $C$.
For example, the noun category applied to a ROUGE alignment would calculate the ROUGE-based precision and recall on the two summaries' nouns.

As long as $C$ is well-understood and interpretable, then so are the category-specific metrics.
Therefore, the precision and recall scores of information-based categories can be used to reason about how much information overlap exists between two summaries.
Similarly, the differences in these metrics between systems can identify whether one system produces better information than the other.

To demonstrate the utility of this evaluation methodology, we calculated the differences in category-specific F$_1$ metrics between the baseline Pointer-Generator \citep{See2017} and state-of-the-art BertSumExtAbs model \citep{Liu2019} on the CNN/DailyMail dataset using ROUGE alignments in Table~\ref{tab:ccndm-comparison}.

Interestingly, some of the largest relative differences between systems are in the information-based categories (e.g., \textsc{vb+nsubj+dobj}).
This is evidence that the state-of-the-art model actually does produce summaries with information that is more similar to the ground-truth.
In contrast, the topic-based metrics (in particular the noun-related categories), show very little relative improvement, suggesting the baseline model discusses the correct topics nearly as well as the state-of-the art model does.

However, this analysis also reveals that the information-based F$_1$ scores are rather low on an absolute scale.
For instance, the \textsc{vb+nsubj} F$_1$ for the BertSumExtAbs model is only 12.0, whereas the \textsc{nn} F$_1$ is 40.9
This large gap indicates that there is significant room for improvement in generating summaries with better information.

Although this evaluation methodology is relatively simple, it provides more insights into the systems' overall behaviors and allows for directly reasoning about the information of the summaries.
The conclusions reached by this evaluation could not have been made by comparing systems using ROUGE or BERTScore alone.

\section{Discussion}
\label{sec:discussion}
\paragraph{Responsiveness Correlations}
Many of the automatic metrics analyzed in this work have demonstrated very high system-level correlations to ground-truth summary responsiveness judgments \citep[Pearson's $r > 0.8$;][]{Dang2008,Dang2009}, so the results that indicate they do not measure information overlap are somewhat surprising.
Since the metrics appear to compare summaries based on the topics they discuss, it is likely that only comparing summary topics is a very strong baseline for these benchmark datasets.

Indeed, we find in Table~\ref{tab:info_vs_topic_correlations} that evaluating with only the NP chunk category-based recall score (which represents little-to-no information) achieves nearly the same correlations as ROUGE on TAC'08.
We hypothesize that further improving correlations to responsiveness judgments will require directly measuring the summaries' information overlap.

\begin{table}[t]
    \centering
    \begin{tabular}{ccccccc}
        \toprule
        & \multicolumn{2}{c}{Summ-Level} & & \multicolumn{2}{c}{Sys-Level} \\
        \cmidrule{2-3} \cmidrule{5-6} 
        Metric & $r$ & $\rho$ & & $r$ & $\rho$ \\
        \midrule
        ROUGE & 0.49 & 0.48 & & 0.80 & 0.80 \\
        \textsc{np-chunks} & 0.45 & 0.44 & & 0.79 & 0.80 \\
        \bottomrule
    \end{tabular}
    \caption{The Pearson $r$ and Spearman $\rho$ correlations of ROUGE and the NP Chunk category-specific recall (Eq. ~\ref{eqn:category_recall}) are very close, demonstrating that a purely topic based comparison (NP chunks) is a very high baseline for content quality correlations on TAC'08.}
    \label{tab:info_vs_topic_correlations}
\end{table}

\paragraph{Limitations}
There are some limitations to our analysis.
First, the results are specific to the datasets and summarization models that were used.
However, TAC'08 and '09 are the benchmark datasets for evaluating content quality and have been widely used to measure the performance of different metrics.
Further, because the results from \S\ref{sec:category_analysis} are consistent across two rather different datasets (TAC and CNN/DailyMail), we believe these results are likely to hold for other datasets as well.

Then, the information categories from \S\ref{sec:category_analysis} do not capture all of the information from a summary.
A phrase like ``the Turkish journalist'' expresses that nationality of the journalist, but this information would not be represented by the tuples included in our analysis.
However, similar representations of information are used in the OpenIE literature \citep{Etzioni2008}, and we do not believe the addition of more information-based categories is likely to significant change the experimental results.
\section{Related Work}
Most of the work that reasons about how to interpret the scores of evaluation metrics does so indirectly through correlations to human judgments \citep{Dang2009,Owczarzak2011}.
However, a high correlation is not conclusive evidence about what a metric measures since it is possible for the metric to directly measure some other aspect of a summary, which is in turn correlated with the ground-truth judgments (see discussion in \S\ref{sec:discussion}).
Our work can be viewed as more direct evidence about what ROUGE and BERTScore measure.

Recent work by \citet{Wang2020} demonstrates that many of the same evaluation metrics covered in this work do not successfully measure the faithfulness of a summary based on low correlations to ground-truth judgments.
The results from our experiments offer an explanation for why this is the case: The metrics do not compare summaries based on their information, therefore they cannot determine if a summary is factually consistent with its input.

Metrics which do attempt to directly measure information overlap between summaries are based on the gold-standard comparison technique, the Pyramid Method \citep{Nenkova2004}.
Although it relies heavily on annotations by experts, there have been attempts to crowdsource \citep{Shapira2019} or automate all or parts of the Pyramid Method \citep{Passonneau2013,Yang2016,Hirao2018} including PyrEval \citep{Gao2019}, which we analyzed in \S\ref{sec:other_metrics}.
These metrics have been met with less success than the text overlap-based ones covered by this work, potentially because measuring information overlap is more difficult than comparing summaries by their topics, and topic-based evaluations strongly correlate to responsiveness judgments (see \S\ref{sec:discussion}).

\section{Conclusion}
In this work, we argued that ROUGE, BERTScore, and many other proposed metrics for evaluating the content quality of summaries largely do not compare summaries based on their information overlap.
The implications of this result are that the summarization community does not have a reliable metric that aligns with its research goal, to generate summaries with high-quality information.
Finally, we proposed a simple and interpretable evaluation methodology which does measure information overlap and provides new insights into model behavior that previous metrics alone could not do.

\bibliography{summarization,emnlp2020,ccg,cited}

\begin{thebibliography}{26}
\expandafter\ifx\csname natexlab\endcsname\relax\def\natexlab#1{#1}\fi

\bibitem[{Dang and Owczarzak(2008)}]{Dang2008}
Hoa~Trang Dang and Karolina Owczarzak. 2008.
\newblock Overview of the tac 2008 update summarization task.
\newblock In \emph{TAC}.

\bibitem[{Dang and Owczarzak(2009)}]{Dang2009}
Hoa~Trang Dang and Karolina Owczarzak. 2009.
\newblock Overview of the tac 2009 summarization track.
\newblock In \emph{proceedings of the Text Analysis Conference}.

\bibitem[{Denkowski and Lavie(2014)}]{Denkowski2014}
Michael~J. Denkowski and Alon Lavie. 2014.
\newblock \href {https://doi.org/10.3115/v1/w14-3348} {Meteor universal:
  Language specific translation evaluation for any target language}.
\newblock In \emph{Proceedings of the Ninth Workshop on Statistical Machine
  Translation, WMT@ACL 2014, June 26-27, 2014, Baltimore, Maryland, {USA}},
  pages 376--380. The Association for Computer Linguistics.

\bibitem[{Etzioni et~al.(2008)Etzioni, Banko, Soderland, and
  Weld}]{Etzioni2008}
Oren Etzioni, Michele Banko, Stephen Soderland, and Daniel~S Weld. 2008.
\newblock Open information extraction from the web.
\newblock \emph{Communications of the ACM}, 51(12):68--74.

\bibitem[{Gao et~al.(2020)Gao, Zhao, and Eger}]{Gao2020}
Yang Gao, Wei Zhao, and Steffen Eger. 2020.
\newblock \href {https://www.aclweb.org/anthology/2020.acl-main.124/} {{SUPERT:
  Towards New Frontiers in Unsupervised Evaluation Metrics for Multi-Document
  Summarization}}.
\newblock In \emph{Proceedings of the 58th Annual Meeting of the Association
  for Computational Linguistics, {ACL} 2020, Online, July 5-10, 2020}, pages
  1347--1354. Association for Computational Linguistics.

\bibitem[{Gao et~al.(2019)Gao, Sun, and Passonneau}]{Gao2019}
Yanjun Gao, Chen Sun, and Rebecca~J. Passonneau. 2019.
\newblock \href {https://doi.org/10.18653/v1/K19-1038} {Automated pyramid
  summarization evaluation}.
\newblock In \emph{Proceedings of the 23rd Conference on Computational Natural
  Language Learning, CoNLL 2019, Hong Kong, China, November 3-4, 2019}, pages
  404--418. Association for Computational Linguistics.

\bibitem[{Giannakopoulos and Karkaletsis(2011)}]{Giannakopoulos2011}
George Giannakopoulos and Vangelis Karkaletsis. 2011.
\newblock \href
  {https://tac.nist.gov/publications/2011/participant.papers/DemokritosGR.proceedings.pdf}
  {Autosummeng and memog in evaluating guided summaries}.
\newblock In \emph{Proceedings of the Fourth Text Analysis Conference, {TAC}
  2011, Gaithersburg, Maryland, USA, November 14-15, 2011}. {NIST}.

\bibitem[{Giannakopoulos and Karkaletsis(2013)}]{Giannakopoulos2013}
George Giannakopoulos and Vangelis Karkaletsis. 2013.
\newblock \href {https://doi.org/10.1007/978-3-642-37256-8\_36} {Summary
  evaluation: Together we stand npower-ed}.
\newblock In \emph{Computational Linguistics and Intelligent Text Processing -
  14th International Conference, CICLing 2013, Samos, Greece, March 24-30,
  2013, Proceedings, Part {II}}, volume 7817 of \emph{Lecture Notes in Computer
  Science}, pages 436--450. Springer.

\bibitem[{Giannakopoulos et~al.(2008)Giannakopoulos, Karkaletsis, Vouros, and
  Stamatopoulos}]{Giannakopoulos2008}
George Giannakopoulos, Vangelis Karkaletsis, George~A. Vouros, and Panagiotis
  Stamatopoulos. 2008.
\newblock \href {https://doi.org/10.1145/1410358.1410359} {Summarization system
  evaluation revisited: N-gram graphs}.
\newblock \emph{{TSLP}}, 5(3):5:1--5:39.

\bibitem[{Hirao et~al.(2018)Hirao, Kamigaito, and Nagata}]{Hirao2018}
Tsutomu Hirao, Hidetaka Kamigaito, and Masaaki Nagata. 2018.
\newblock \href {https://doi.org/10.18653/v1/d18-1450} {Automatic pyramid
  evaluation exploiting edu-based extractive reference summaries}.
\newblock In \emph{Proceedings of the 2018 Conference on Empirical Methods in
  Natural Language Processing, Brussels, Belgium, October 31 - November 4,
  2018}, pages 4177--4186. Association for Computational Linguistics.

\bibitem[{Hovy et~al.(2006)Hovy, Lin, Zhou, and Fukumoto}]{Hovy2006}
Eduard~H. Hovy, Chin{-}Yew Lin, Liang Zhou, and Junichi Fukumoto. 2006.
\newblock \href
  {http://www.lrec-conf.org/proceedings/lrec2006/summaries/438.html} {Automated
  summarization evaluation with basic elements}.
\newblock In \emph{Proceedings of the Fifth International Conference on
  Language Resources and Evaluation, {LREC} 2006, Genoa, Italy, May 22-28,
  2006}, pages 899--902. European Language Resources Association {(ELRA)}.

\bibitem[{Lin(2004)}]{Lin2004}
Chin-Yew Lin. 2004.
\newblock \href {https://www.aclweb.org/anthology/W04-1013} {{ROUGE}: A package
  for automatic evaluation of summaries}.
\newblock In \emph{Text Summarization Branches Out}, pages 74--81, Barcelona,
  Spain. Association for Computational Linguistics.

\bibitem[{Liu and Lapata(2019)}]{Liu2019}
Yang Liu and Mirella Lapata. 2019.
\newblock \href {https://doi.org/10.18653/v1/D19-1387} {Text summarization with
  pretrained encoders}.
\newblock In \emph{Proceedings of the 2019 Conference on Empirical Methods in
  Natural Language Processing and the 9th International Joint Conference on
  Natural Language Processing, {EMNLP-IJCNLP} 2019, Hong Kong, China, November
  3-7, 2019}, pages 3728--3738. Association for Computational Linguistics.

\bibitem[{Louis and Nenkova(2013)}]{Louis2013}
Annie Louis and Ani Nenkova. 2013.
\newblock \href {https://doi.org/10.1162/COLI\_a\_00123} {Automatically
  assessing machine summary content without a gold standard}.
\newblock \emph{Computational Linguistics}, 39(2):267--300.

\bibitem[{Nallapati et~al.(2016)Nallapati, Zhou, dos Santos,
  G{\"{u}}l{\c{c}}ehre, and Xiang}]{Nallapati2016}
Ramesh Nallapati, Bowen Zhou, C{\'{\i}}cero~Nogueira dos Santos, {\c{C}}aglar
  G{\"{u}}l{\c{c}}ehre, and Bing Xiang. 2016.
\newblock \href {https://doi.org/10.18653/v1/k16-1028} {Abstractive text
  summarization using sequence-to-sequence rnns and beyond}.
\newblock In \emph{Proceedings of the 20th {SIGNLL} Conference on Computational
  Natural Language Learning, CoNLL 2016, Berlin, Germany, August 11-12, 2016},
  pages 280--290. {ACL}.

\bibitem[{Nenkova and Passonneau(2004)}]{Nenkova2004}
Ani Nenkova and Rebecca~J. Passonneau. 2004.
\newblock \href {https://www.aclweb.org/anthology/N04-1019/} {Evaluating
  content selection in summarization: The pyramid method}.
\newblock In \emph{Human Language Technology Conference of the North American
  Chapter of the Association for Computational Linguistics, {HLT-NAACL} 2004,
  Boston, Massachusetts, USA, May 2-7, 2004}, pages 145--152. The Association
  for Computational Linguistics.

\bibitem[{Owczarzak and Dang(2011)}]{Owczarzak2011}
Karolina Owczarzak and Hoa~Trang Dang. 2011.
\newblock {Overview of the TAC 2011 summarization track: Guided task and AESOP
  task}.
\newblock In \emph{Proceedings of the Text Analysis Conference (TAC 2011),
  Gaithersburg, Maryland, USA, November}.

\bibitem[{Passonneau et~al.(2013)Passonneau, Chen, Guo, and
  Perin}]{Passonneau2013}
Rebecca~J. Passonneau, Emily Chen, Weiwei Guo, and Dolores Perin. 2013.
\newblock \href {https://www.aclweb.org/anthology/P13-2026/} {Automated pyramid
  scoring of summaries using distributional semantics}.
\newblock In \emph{Proceedings of the 51st Annual Meeting of the Association
  for Computational Linguistics, {ACL} 2013, 4-9 August 2013, Sofia, Bulgaria,
  Volume 2: Short Papers}, pages 143--147. The Association for Computer
  Linguistics.

\bibitem[{Peyrard and Eckle{-}Kohler(2017)}]{Peyrard2017}
Maxime Peyrard and Judith Eckle{-}Kohler. 2017.
\newblock \href {https://doi.org/10.18653/v1/P17-1100} {Supervised learning of
  automatic pyramid for optimization-based multi-document summarization}.
\newblock In \emph{Proceedings of the 55th Annual Meeting of the Association
  for Computational Linguistics, {ACL} 2017, Vancouver, Canada, July 30 -
  August 4, Volume 1: Long Papers}, pages 1084--1094. Association for
  Computational Linguistics.

\bibitem[{See et~al.(2017)See, Liu, and Manning}]{See2017}
Abigail See, Peter~J. Liu, and Christopher~D. Manning. 2017.
\newblock \href {https://doi.org/10.18653/v1/P17-1099} {Get to the point:
  Summarization with pointer-generator networks}.
\newblock In \emph{Proceedings of the 55th Annual Meeting of the Association
  for Computational Linguistics, {ACL} 2017, Vancouver, Canada, July 30 -
  August 4, Volume 1: Long Papers}, pages 1073--1083. Association for
  Computational Linguistics.

\bibitem[{Shapira et~al.(2019)Shapira, Gabay, Gao, Ronen, Pasunuru, Bansal,
  Amsterdamer, and Dagan}]{Shapira2019}
Ori Shapira, David Gabay, Yang Gao, Hadar Ronen, Ramakanth Pasunuru, Mohit
  Bansal, Yael Amsterdamer, and Ido Dagan. 2019.
\newblock \href {https://doi.org/10.18653/v1/n19-1072} {Crowdsourcing
  lightweight pyramids for manual summary evaluation}.
\newblock In \emph{Proceedings of the 2019 Conference of the North American
  Chapter of the Association for Computational Linguistics: Human Language
  Technologies, {NAACL-HLT} 2019, Minneapolis, MN, USA, June 2-7, 2019, Volume
  1 (Long and Short Papers)}, pages 682--687. Association for Computational
  Linguistics.

\bibitem[{Tratz and Hovy(2008)}]{Tratz2008}
Stephen Tratz and Eduard~H. Hovy. 2008.
\newblock \href
  {https://tac.nist.gov/publications/2008/additional.papers/ISI.proceedings.pdf}
  {Summarization evaluation using transformed basic elements}.
\newblock In \emph{Proceedings of the First Text Analysis Conference, {TAC}
  2008, Gaithersburg, Maryland, USA, November 17-19, 2008}. {NIST}.

\bibitem[{Wang et~al.(2020)Wang, Cho, and Lewis}]{Wang2020}
Alex Wang, Kyunghyun Cho, and Mike Lewis. 2020.
\newblock \href {https://www.aclweb.org/anthology/2020.acl-main.450/} {Asking
  and answering questions to evaluate the factual consistency of summaries}.
\newblock In \emph{Proceedings of the 58th Annual Meeting of the Association
  for Computational Linguistics, {ACL} 2020, Online, July 5-10, 2020}, pages
  5008--5020. Association for Computational Linguistics.

\bibitem[{Yang et~al.(2016)Yang, Passonneau, and de~Melo}]{Yang2016}
Qian Yang, Rebecca~J. Passonneau, and Gerard de~Melo. 2016.
\newblock \href
  {http://www.aaai.org/ocs/index.php/AAAI/AAAI16/paper/view/12481} {{PEAK:}
  pyramid evaluation via automated knowledge extraction}.
\newblock In \emph{Proceedings of the Thirtieth {AAAI} Conference on Artificial
  Intelligence, February 12-17, 2016, Phoenix, Arizona, {USA}}, pages
  2673--2680. {AAAI} Press.

\bibitem[{Zhang et~al.(2019)Zhang, Kishore, Wu, Weinberger, and
  Artzi}]{Zhang2019}
Tianyi Zhang, Varsha Kishore, Felix Wu, Kilian~Q. Weinberger, and Yoav Artzi.
  2019.
\newblock \href {http://arxiv.org/abs/1904.09675} {Bertscore: Evaluating text
  generation with {BERT}}.
\newblock \emph{CoRR}, abs/1904.09675.

\bibitem[{Zhao et~al.(2019)Zhao, Peyrard, Liu, Gao, Meyer, and Eger}]{Zhao2019}
Wei Zhao, Maxime Peyrard, Fei Liu, Yang Gao, Christian~M. Meyer, and Steffen
  Eger. 2019.
\newblock \href {https://doi.org/10.18653/v1/D19-1053} {Moverscore: Text
  generation evaluating with contextualized embeddings and earth mover
  distance}.
\newblock In \emph{Proceedings of the 2019 Conference on Empirical Methods in
  Natural Language Processing and the 9th International Joint Conference on
  Natural Language Processing, {EMNLP-IJCNLP} 2019, Hong Kong, China, November
  3-7, 2019}, pages 563--578. Association for Computational Linguistics.

\end{thebibliography}
\bibliographystyle{acl_natbib}

\end{document}